# Forecasting Solar Power Generation on the basis of Predictive and Corrective Maintenance Activities

Soham Vyas

Department of Computer Science & Engineering, PDEU, Gandhinagar, India, Soham.vmtds21@sot.pdpu.ac.in

Sanskar Bhuwania

Department of Computer Science & Engineering, PDEU, Gandhinagar,

sanskar.bce18@sot.pdpu.ac.in

Brijesh Tripathi

Department of Solar Energy, PDEU, Gandhinagar, India, brijesh.tripathi@sot.pdpu.ac.in

Yuvraj Goyal

Department of Computer Science & Engineering, PDEU, Gandhinagar,

India, Yuvraj.gmtds21@sot.pdpu.ac.in

Hardik Patel

Department of Information and Communication Technology, PDEU, Gandhinagar, India, Hardik.patel@sot.pdpu.ac.in

Neel Bhatt

Department of Computer Science & Engineering, PDEU, Gandhinagar,

India, Neel.bmtds20@sot.pdpu.ac.in

Shakti Mishra

Department of Computer Science & Engineering, PDEU, Gandhinagar, India, Shakti.mishra@sot.pdpu.ac.in

*Abstract*—Solar energy forecasting has seen tremendous growth in the last decade using historical time series collected from a weather station, such as weather variables wind speed and direction, solar irradiance, and temperature. It helps in the overall management of solar power plants. However, the solar power plant regularly requires preventive and corrective maintenance activities that further impact energy production. This paper presents a novel work for forecasting solar power energy production based on maintenance activities, problems observed at a power plant, and weather data. The results accomplished on the dataset obtained from the 1MW solar power plant of PDEU (our university) that has generated data set with 13 columns as daily entries from 2012 to 2020. There are 12 structured columns and one unstructured column with manual text entries about different maintenance activities, problems observed, and weather conditions daily. The unstructured column is used to create a new feature column vector using Hash Map, flag words, and stop words. The final dataset comprises five important feature vector columns based on correlation and causality analysis.

Further, the random forest regression is used to compute the impact of maintenance activities on the total energy output. The causality and correlation analysis has shown that the five feature vectors are interdependent time series variables. Next, Vector Autoregression (VAR) is chosen for simultaneous forecasting of total power generation for 3, 5, 7, 10, 12, and 30 days ahead using the VAR model. The results have shown that the root means square percentage error (RMSPE) in total power generation forecasting is less than 10% for different days. This research has proven that the spikes in total power generation forecasting can be traced and tracked better using daily maintenance activities, observed problems, and weather conditions.

I. INTRODUCTION

Solar power generation has the potential to mitigate climate change by reducing the carbon footprint. It has had better market penetration in recent years because of awareness about clean and green energy and its affordable cost. Solar power plants require various planned and unplanned maintenance activities for better energy output. These maintenance activities include PV module cleaning and maintenance, PV module positioning in the field, inverter maintenance, etc. Solar energy forecasting is usually done using past time series data acquired from weather stations such as wind pressure, humidity, temperature, satellite imagery, etc. In this research work, total solar power generation forecasting is proposed by using different maintenance activities, problems observed, and weather data. Next, we have carried out a literature survey to understand the contemporary work done in this area.

Fuzzy logic, AI models, and genetic algorithms are used to predict and model solar radiation, seizing, performances, and controls of the solar photovoltaic (PV) systems in [1]. Ensemble of deep ConvNets is proposed for multistep solar forecasting without additional time series models like RNN or LSTM and exogenous variables in [2] with 22.5% RMSE. Mycielski-Markov is utilized to forecast solar power generation for a short period in [3] with 32.65% RMSE. Feedforward neural network-based solar irradiance prediction is followed by LSTM-based solar power generation prediction for a short period [4] with 98.70 average RMSE. The ensemble approach is proposed based on long short-term memory (LSTM), gated recurrent unit (GRU), Autoencoder LSTM (Auto-LSTM), and Auto-GRU for solar power generation forecasting in [5] without considering any maintenance activities. Generic fault/status prediction and specific fault prediction by unsupervised clustering and neural network by using data of 10MW solar power plant and one hundred inverters of three different technology brands [6]. This model can predict generic faults up to 7 days in advance with 95% sensitivity and specific defects before some hours to 7 days [6]. Intra hour, short term, medium term, long term, ramp forecasting, and load forecasting are proposed for renewable EnergyEnergy like wind and solar EnergyEnergy [7]. Solar power generation is reduced by 17.4% per month because of dust on solar collectors [8]. Day-ahead forecasting of 1MW solar power plant output is proposed in the American Southwest with 10.3% to 14% RMSE [9]. Solar power generation is forecast using different neural network models like LSTM, MLP, LRNN, feedforward, ARMA, ARIMA, SARIMA, and 3640 hours of data for a 20MW power plant [10]. Six-hour-ahead solar power forecasting is proposed using an autoregressive forecasting model at residential and medium voltage substation levels [11]. The autoregressive model of [11] claims 8% to 10% improvements in results. Two-stage probabilistic solar power forecasting is proposed in [12], the first stage is used to predict solar irradiance, and the second stage is used to predict solar power. The model of [12] results in minimum loss and the highest daily profit in the energy market. A robust auto encoder-gated recurrent unit (AE-GRU) model is used to forecast solar power generation for 24 h, 48 h, and 15 days [13]. Sparsity promoting LASSO-VAR structures are proposed and fitted with alternating direction method of multipliers (ADMM), 1hour and 15-minute resolution for solar power forecasting in [14]. The LASSO-VAR model of [14] improves 11% in the forecasting. The probabilistic solar power forecasting is proposed and compared with the autoregressive method in [15], which results in RMSE of 8% to 12%. A nonlinear autoregressive neural network with an exogenous input model is proposed with Levenberg-Marquardt, Bayesian regularization, scaled conjugate gradient, and Broyden-Fletcher-Goldfarb-Shanno (BFGS) algorithms for solar power forecasting over NIGERIA [16]. The models of [16] result in RMSE values ranging from 0.162 to 0.544 W/m$^2$. Five-minute-ahead forecasts are produced and evaluated using point and probabilistic forecast skill scores and calibration using sparse vector autoregression for 22 wind farms in Australia [17]. The LASSO vector autoregression model is proposed for very short-term wind power forecasting [18]. A vector autoregression weather model is proposed for electricity supply and demand modeling with six hours ahead forecasting with less RMSE [19]. Graph-convolutional long short-term memory (GCLSTM) and the graph-convolutional transformer (GCTrafo), named two novel graph neural network models, are proposed for multi-site photovoltaic power forecasting with 12.6% and 13.6% NRMSE respectively [20]. The following sections are data set preprocessing, methodology, results and analysis, conclusions, and future work.

## II. Data set preprocessing

Pandit Deendayal Energy University (PDEU), Gandhinagar, and Gujarat Energy Research and Management Institute (GERMI) set up a 1 MW Solar Power plant in 2012. The dataset obtained from this solar power plant has been used for this work from 2012-to 2020. This dataset has daily entries of 13 columns from 2012 to 2020. The solar plant consists of five sets of PV modules. Three out of these five sets are "poly-crystalline" based, and each has the capacity of approximately 250KW. The remaining two PV modules are "thin-film amorphous silicon" and "Concentrate Photovoltaic" based with capacities of approximately 250 KW and 15 KW, respectively. There are four sets of PV modules, and each set has approximately a 250KW capacity. The fifth set of PV modules has approximately a 15KW capacity. The dataset has five columns for power generation from five sets of PV modules and the other columns are "date", "Total power generation (KWH)", "aggregate meter reading (KWH)", "difference", "Seeds data (KWH)", "insolation", "PR (%)" and "any issues/problems observed". As discussed above, there are 13 columns in this data set, and it is semi-structured because the last column, "any issues/problems observed," has text data that includes day-wise manually entered weather information, maintenance issues, grid failure, module cleaning information, etc. from 2012 to 2020. The first and most important research challenge is to create the different features from the last column, "any issues/problems observed ."This research challenge was addressed by creating a nested hash map with different rules. The key contains the possible feature label as a text, and the value is a 2-dimensional array. One array has words representing the maintenance issues, a problem observed, or weather conditions. The second array has to stop words that prevent overlapping and duplication of the maintenance issues or problems observed or weather conditions. Each key is the new feature (maintenance or problem observed or weather condition) column, and the value is tokenized as one of the new features is present on a particular day. New feature vectors are created with labels from the column "any issues/problems observed ."Now, each new feature column vector label value one is replaced by its percentage of the occurrence. New feature vectors created are "Grid Failure", "Inverter Failure", "Module Cleaning", "Rainy Day", "No Module Cleaning", "Transformer Replacement and Maintenance", "Cable and Fuse Maintenance", "Plant Shutdown,", "Internet", "Battery", "Cloudy day", "Module Cleaning by Rain" by using the above approach. There are only five columns, "Total generation (KWH)", "Grid Failure", "Inverter Failure", "Module Cleaning", and "Cloudy," in the final dataset based on the correlation and causality analysis. Vector autoregression (VAR) model is selected for simultaneous forecasting of total power generation and new features because they are inter-dependant time series data.

## III. Methodology

In this paper, solar power generation is forecasted using maintenance activities. It is novel work, and there is not much research done on this topic. The power generation prediction is formulated as a regression problem to understand the usage of maintenance issues. The labels of processed datasets have been used to feed the regression model, and the future maintenance variable has been considered test data. Random Forest Regression is applied to this data set, and it has been observed that the maintenance issues can be used as variables to forecast the power generation.

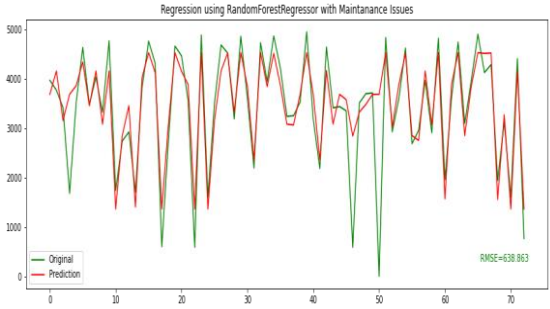

Fig. 1 Regression Model using Random Forest Regressor

*Vector Autoregression*

VAR models are used for multivariate time series. The VAR models consider each variable as a linear function of past lags of itself and past lags of the other variables. Five variables <Total Generation (KWH), Grid failure, Inverter Failure, Module Cleaning, and Cloudy > have been considered and modeled as a system of equations with one equation per variable in time series. Let us consider if we have two variables (Time series), $Y_1$ and $Y_2$, and we need to forecast the values of these variables at a time (t). To calculate $Y_1$ (t), VAR will use the past values of both $Y_1$ and $Y_2$. Likewise, to compute $Y_2$ (t), the past values of both $Y_1$ and $Y_2$ are used. For example, the system of equations for a VAR model with two-time series (variables `$Y_1$` and `$Y_2$`) is as follows:

$$Y_{1,t} = \alpha_1 + \beta_{11,1}Y_{1,t-1} + \beta_{12,1}Y_{2,t-1} + \varepsilon_{1,t}$$
$$Y_{2,t} = \alpha_2 + \beta_{21,1}Y_{1,t-1} + \beta_{22,1}Y_{2,t-1} + \varepsilon_{2,t} \quad (1)$$

The vector autoregressive model of order one is denoted as VAR (1). Similarly, in a VAR (2) model, the lag two values for all variables are added to the right sides of the equations. In the case of five Y-variables (or time series), there would be ten predictors on the right side of each equation, five lag one term and five lag two terms. For a VAR (p) model, the first p lags of each variable in the system would be used as regression predictors for each variable. As per equation (1), the data follows stationarity and the causality test. In a causality test, the data follows the interconnected time series dependencies. Akaike information criterion (AIC) is the model $M_k$ with dimension k is defined as where L ($M_k$) is the likelihood corresponding to the model $M_k$. The first term in AIC is twice the negative log-likelihood, which turns out to be the residual sum of squares corresponding to the model $M_k$ for the linear regression model with a Gaussian likelihood [21]. AIC has been computed using data before forecasting, and optimal AIC was derived for the specific lag days to fit the VAR. After checking the model with acquired lag day, the coefficient matrix is computed for each equation. Here five variables are used for the endogenous attribute. Forecasting results with multiple periods of days will give us an understanding of how power generation varies and the probability of the various spikes. Our equation has coefficients for years and tries to forecast with the help of lag days. It will become our return value, and the result will be separated into multiple scenarios.

## IV. RESULTS AND ANALYSIS

The VAR model is given to understand the effect of different days on total power generation forecasting results for different days.

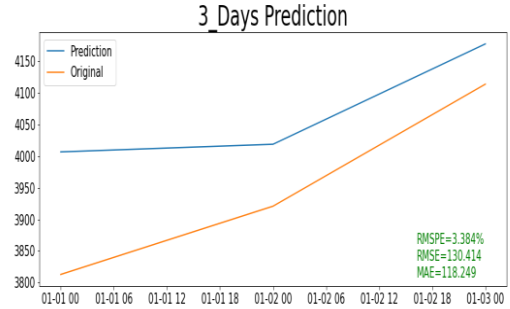

Fig.2 Three days ahead of total power generation forecasting by using the VAR model

As shown in fig.2, the RMSPE, RMSE, and MAE are 3.38%, 130.414, and 118.249, respectively, for three days ahead of total power generation forecasting.

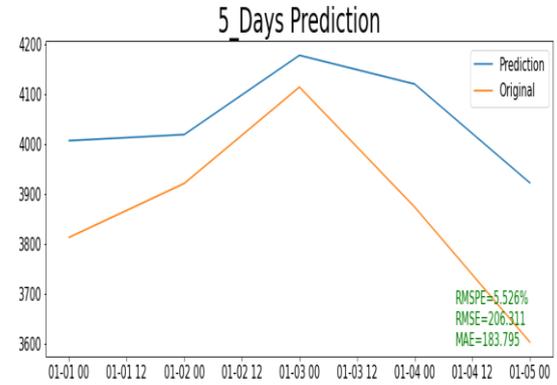

Fig.3 Five days ahead total power generation forecasting by using the VAR model

As shown in fig.3, the RMSPE, RMSE, and MAE are 5.52%, 206.331, and 183.795, respectively, for five days ahead of total power generation forecasting.

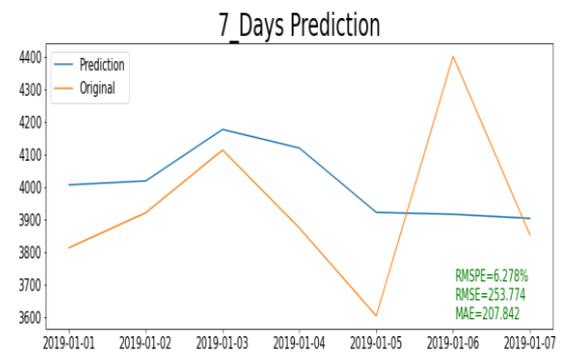

Fig.4 Seven days ahead total power generation forecasting by using the VAR model

As shown in fig.4, the RMSPE, RMSE, and MAE are 6.27%, 253.774, and 207.842, respectively, for seven days ahead of total power generation forecasting.

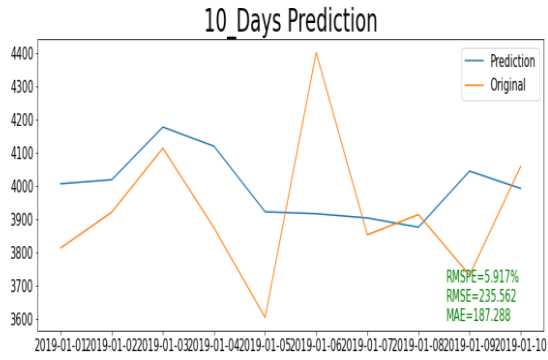

Fig.5 Ten days ahead total power generation forecasting by using the VAR model

As shown in fig.5, the RMSPE, RMSE, and MAE are 5.91%, 235.562, and 187.288, respectively, for ten days ahead of total power generation forecasting.

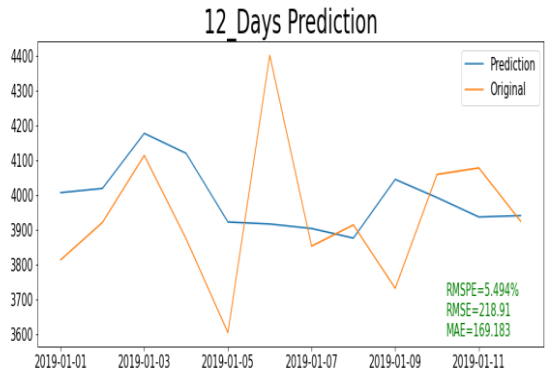

Fig.6 Twelve days ahead total power generation forecasting by using the VAR model

As shown in fig.6, the RMSPE, RMSE, and MAE are 5.49%, 218.91, and 169.183, respectively, for 12 days ahead of total power generation forecasting.

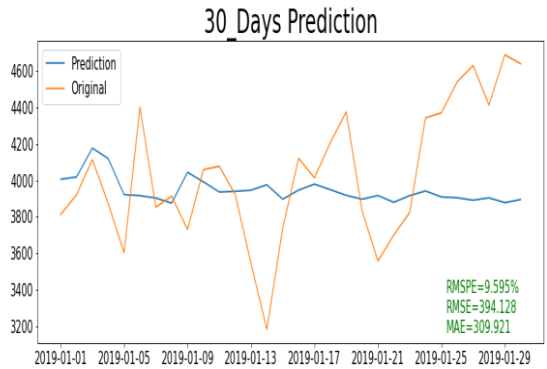

Fig.7 Thirty days ahead total power generation forecasting by using the VAR model

As shown in fig.7, the RMSPE, RMSE, and MAE are 9.59%, 394.128, and 309.921, respectively, for 30 days ahead of total power generation forecasting.

TABLE I. RMSPE, RMSE, AND MAE IN TOTAL POWER GENERATION FORECASTING FOR DIFFERENT DAYS.

| Days | RMSPE | RMSE | MAE |
|---|---|---|---|
| 3 | 3.38 | 130.414 | 118.249 |
| 5 | 5.52 | 206.331 | 183.795 |
| 7 | 6.27 | 253.744 | 207.842 |
| 10 | 5.91 | 235.562 | 187.288 |
| 12 | 5.49 | 218.91 | 169.183 |
| 30 | 9.59 | 394.128 | 309.921 |

As shown by figures 2 to 7, the VAR model can predict almost all power generation spikes, which is the most crucial point of this research. The total power generation spikes are due to different maintenance activities, problems, and weather conditions. The real power generation forecasting error is lowest three days ahead of forecasting. Table I shows that the error in total power generation forecasting is less than 10% for different days. VAR model can forecast all the new features "Grid Failure", "Inverter Failure", "Module Cleaning", "Rainy Day", "No Module Cleaning", "Transformer Replacement," and "Maintenance", "Cable and Fuse Maintenance", "Plant Shutdown", "Internet", "Battery", "Cloudy day", "Module Cleaning by Rain" and "total power generation" because all these are interdependent time series.

## V. CONCLUSIONS

The research work in the paper presents the forecasting of total power generation based on various maintenance activities carried out in solar power plants. Scheduled maintenance activities in the power plant impact energy production. This work involves transforming the unstructured dataset into structured form with twelve new feature vectors using HashMap, flag words, and stop words. Further, Random Forest Regressor is used to analyze the impact of maintenance activities on forecasting total power generation. The same outcome has shown that the total power generation prediction is perfect because of the maintenance activities. The maintenance activities are not available for forecasting, so maintenance activities should be predicted before the total power generation forecasting. Vector Auto Regression-based model is used for forecasting multivariate time-series considering five variables "Total Power Generation (KWH)", "Grid Failure", "Inverter Failure", "Module Cleaning", and "Cloudy". VAR can forecast total power generation along with forecasting four maintenance activities. The three days ahead total power generation forecasting has the lowest error compared to other results. Total power generation forecasting is implemented in two stages in the literature review. The first stage predicts solar irradiance or maintenance activities, problems, and weather conditions, and the second stage is total power generation forecasting. In this research work, forecasting of solar power generation and maintenance activities, problems, and weather conditions are all done simultaneously.

## VI. FUTURE WORK

In the future, this work shall be extended by comparing the total power generation forecasting using different models and the inclusion of two essential feature vectors, "solar irradiance" and "insolation," in the current VAR model for solar power generation forecasting. It is also planned to forecast the solar power generation for an individual set of

PV modules to determine the impact of different PV modules on the forecasting. The evaluation shall be based on the overall effect of varying PV modules, "solar irradiance", "insolation," and daily maintenance activities, problems observed, and weather conditions on the total power generation forecasting.